\documentclass[10pt, twocolumn]{article}
\usepackage[a4paper, margin=1.5cm]{geometry}
\usepackage{chemformula} % Formula subscripts using \ch{}
\usepackage[T1]{fontenc} % Use modern font encodings
\usepackage{amsmath}
\usepackage{amssymb}
\usepackage{amsfonts}
\usepackage{graphicx}
\usepackage[labelfont=bf]{caption}
\usepackage{subcaption}
\usepackage{bbm}
\usepackage{booktabs}
\usepackage{multirow}
\usepackage{longtable}
\usepackage{makecell}
\usepackage{adjustbox}
\usepackage{authblk}
\usepackage{placeins}
\usepackage[superscript]{cite}
\usepackage{url}
\usepackage{tcolorbox} % for boxes

\title{\textbf{Drug discovery with explainable artificial intelligence}}

\author[+]{Jos\'e Jim\'enez-Luna}
\author[+]{Francesca Grisoni}
\author[*]{Gisbert Schneider}

\affil[ ]{\normalsize{ETH Zurich, Department of Chemistry and Applied Biosciences, RETHINK, Vladimir-Prelog-Weg 4, 8093 Zurich, Switzerland.}}
\affil[*]{\normalsize{Correspondence: gisbert@ethz.ch}}
\affil[+]{\small{These authors contributed equally to this work}}

\date{}

\begin{document}

\maketitle

\begin{abstract}

Deep learning bears promise for drug discovery, including advanced image analysis, prediction of molecular structure and function, and automated generation of innovative chemical entities with bespoke properties. Despite the growing number of successful prospective applications, the underlying mathematical models often remain elusive to interpretation by the human mind. There is a demand for `explainable' deep learning methods to address the need for a new narrative of the machine language of the molecular sciences. This review summarizes the most prominent algorithmic concepts of explainable artificial intelligence, and dares a forecast of the future opportunities, potential applications, and remaining challenges.

\end{abstract}

\noindent\rule[0.5ex]{\linewidth}{1pt}

\section{Introduction}

Several concepts of `artificial intelligence' (AI) have been adopted for computer-assisted drug discovery \cite{GawehnMolInf2016, zhang2017machine, CHEN20181241, tang2018deep, yang2019concepts, muratov2020qsar, hemmerich2020silico}. Deep learning algorithms, \textit{i.e.}, artificial neural networks with multiple processing layers \cite{lecun2015deep}, are currently receiving particular attention, owing to their capacity to model complex nonlinear input-output relationships, and perform pattern recognition and feature extraction from low-level data representations \cite{lecun2015deep, schmidhuber2015deep}.

Certain deep learning models have been shown to match or even exceed the performance of the existing machine-learning and quantitative structure-activity relationship (QSAR) methods for drug discovery \cite{lenselink2017beyond, goh2017chemception, unterthiner2014deep, wallach2015atomnet, muratov2020qsar}. Moreover, deep learning has boosted the potential and broadened the applicability of computer-assisted discovery, \textit{e.g.,} for molecular de novo design \cite{mendez2020novo, merk2018novo, zhavoronkov2019deep}, chemical synthesis planning \cite{schwaller2018found, coley2018machine, coley2019graph}, protein structure prediction \cite{senior2020improved, yang2020improved}, and macromolecular target identification \cite{ozturk2018deepdta, zeng2020target, jimenez2018pathwaymap}. The ability to capture intricate nonlinear relationships between input data (\textit{e.g.}, chemical structure representations) and the associated output (\textit{e.g.}, assay readout) often comes at the price of limited comprehensibility of the resulting model. While there have been efforts to explain QSARs in terms of algorithmic insights and molecular descriptor analysis \cite{robinson2017rf, webb2014feature, grisoni2019compara, polishchuk2016struct, kuz2011interpretation, chen2019np, rosenbaum2011interpreting, riniker2013similarity, grisoni2016investigating, todeschini1994new}, deep neural network models notoriously elude immediate accessibility by the human mind \cite{rudin2019stop, schneider2019mind}.

In an effort to mitigate the lack of interpretability of deep learning models, attention has been drawn to explainable artificial intelligence (XAI) \cite{lipton2018mythos, murdoch2019interpretable}. Providing informative explanations of AI models aims to (i) render the underlying decision-making process transparent (`understandable') \cite{doshi2017towards}, (ii) avoid correct predictions for the wrong reasons (the so-called `clever Hans effect' \cite{lapuschkin2019cleverhans}), (iii) avert unfair biases or unethical discrimination \cite{miller2019explanation}, and (iv) bridge the gap between the machine learning community and other scientific disciplines. In medicinal chemistry, XAI already enables the mechanistic interpretation of drug action \cite{xu_2017_acute_oral, ciallella_2019}, and contributes to drug safety enhancement \cite{dey2018predicting} and organic synthesis planning \cite{schwaller2018found, yan2020interpretable}. Intensified efforts towards interpreting deep learning models will help increase their reliability and foster their acceptance and usage in drug discovery and medicinal chemistry projects \cite{sheridan2019interpretation, manica2019toward, vzuvela2018interpretation, preuer2019interpretable}. The availability of `rule of thumb' scores correlating biological effects with physicochemical properties \cite{gupta2019blood, rankovic2017cns, wager2016central, ritchie2013increasing, lesson2015molecular} underscores the willingness, in certain situations, to sacrifice accuracy in favour of models that better fit human intuition. Thus, blurring the lines between the `two QSARs' \cite{fujita2016} (\textit{i.e.}, mechanistically interpretable \textit{vs.} highly accurate models) can be the key for accelerated drug discovery \cite{schneider2019rethinking}.

While the exact definition of XAI is still under debate \cite{guidotti2018survey, miller2019explanation}, several aspects of XAI are certainly desirable in drug design applications \cite{lipton2018mythos}: (i) \textit{transparency}, \textit{i.e.,} knowing how the system reached a particular answer, (ii) \textit{justification}, \textit{i.e.,} elucidating why the answer provided by the model is acceptable, (iii) \textit{informativeness}, \textit{i.e.,} providing new information to human decision-makers, and (iv) \textit{uncertainty estimation}, \textit{i.e.,} the quantification of how reliable a prediction is. Moreover, AI model explanations can be \textit{global} (\textit{i.e.}, summarizing the relevance of input features in the model) or \textit{local} (\textit{i.e.} on individual predictions) \cite{lundberg2020local}. Furthermore, XAI can be model-dependent or agnostic, which in turn affects the potential applicability of each method.

The field of XAI is still in its infancy but moving forward at a fast pace, and we expect an increase of its relevance in the years to come. In this review, we aim to provide a comprehensive overview of recent XAI research, including its benefits, limitations and future opportunities for drug discovery. In what follows, after providing an introduction to the most relevant XAI methods structured into conceptual categories, the existing and some of the potential applications to drug discovery are highlighted. Finally, we discuss the limitations of contemporary XAI and point to the potential methodological improvements needed to foster practical applicability of these techniques to pharmaceutical research. 

\begin{table*}[t!]
\centering
\caption{Computational approaches toward explainable AI, categorized according to the respective methodological concept. For each family of approaches, a brief description of its aim is provided, its specific methods, and reported applications in drug discovery. `\textit{Not reported}' refers to families of methods that, to the best of our knowledge, have not been yet applied in drug discovery. Potential applications of these are discussed in the main text.}
\label{tab:summary}
\renewcommand\extrarowheight{20pt}  %% modify cell height here
\resizebox{\textwidth}{!}{%
\begin{tabular}{@{}llll@{}}
\toprule
\\[-4\normalbaselineskip] %% remove row height from header
\multicolumn{1}{l}{\textbf{Family}} &
  \multicolumn{1}{l}{\textbf{Aim}} &
  \multicolumn{1}{l}{\textbf{Methods}} &
  \multicolumn{1}{l}{\textbf{Reported applications in drug discovery}} \\
\midrule
  \\[-4\normalbaselineskip]
\multirow[t]{3}{*}{Feature attribution} &
  \multirow[t]{3}{0.3\textwidth}{Determine local feature importance towards a prediction.} &
  - Gradient-based &
  \multirow[t]{3}{0.4\textwidth}{Ligand pharmacophore identification \cite{mccloskey2019using, rodriguez2019interpretation, rodriguez2020interpretation}, structural alerts for adverse effect \cite{pope2019explainability}, protein-ligand interaction profiling \cite{hochuli2018visualizing}.}\\
 &
   &
  - Surrogate models &
   \\
 &
   &
  - Perturbation-based &
   \\
  \Xhline{0.1\arrayrulewidth}
  \\[-4\normalbaselineskip]
\multirow[t]{3}{*}{Instance-based} &
  \multirow[t]{3}{0.3\textwidth}{Compute a subset of features that need to be present or absent to guarantee or change a prediction.} &
  - Anchors &
  \multirow[t]{3}{0.4\textwidth}{\textit{Not reported}} \\
 &
   &
  - Counterfactual instances &
   \\
 &
   &
  - Contrastive explanations &
   \\
    \Xhline{0.1\arrayrulewidth}
  \\[-4\normalbaselineskip]
\multirow[t]{2}{*}{Graph-convolution-based} &
  \multirow[t]{2}{0.3\textwidth}{Interpret models within the message-passing framework.} &
  - Sub-graph approaches &
  \multirow[t]{5}{0.4\textwidth}{Retrosynthesis elucidation \cite{ishida2019prediction}, toxicophore and pharmacophore identification \cite{preuer2019interpretable}, ADMET \cite{shang2018edge, ryu2018deeply},  and reactivity prediction \cite{coley2019graph}.} \\
 &
   &
  - Attention-based &
   \\
    \Xhline{0.1\arrayrulewidth}
  \\[-4\normalbaselineskip]
\multirow[t]{4}{*}{Self-explaining} &
  \multirow[t]{4}{0.3\textwidth}{Develop models that are explainable by design.} &
  - Prototype-based &
  \multirow[t]{4}{0.4\textwidth}{\textit{\textit{Not reported}}} \\
 &
   &
 - Self-explaining neural networks &
   \\
 &
   &
  - Concept learning &
   \\
 &
   &
  - Natural language explanations &
   \\
   
   \Xhline{0.1\arrayrulewidth}
  \\[-4\normalbaselineskip]
\multirow[t]{3}{*}{Uncertainty estimation} &
  \multirow[t]{3}{0.3\textwidth}{Quantify the reliability of a prediction.} &
  - Ensemble-based &
  \multirow[t]{3}{0.4\textwidth}{Reaction prediction \cite{schwaller2019molecular}, active learning \cite{zhang2019bayesian}, molecular activity prediction \cite{liu2019ad}.} \\
 &
   &
 - Probabilistic &
   \\
 &
   &
  - Other approaches &
   \\
   \Xhline{0.1\arrayrulewidth}

\end{tabular}%
}

\end{table*}

\section{State of the art}

This section aims to provide a concise overview of modern XAI approaches, and exemplify their use in computer vision, natural-language processing, and discrete mathematics. We then highlight selected case studies in drug discovery and propose potential future areas of XAI research. A summary of the methodologies and their goals, along with reported applications is provided in Table~\ref{tab:summary}. In what follows, without loss of generality, $f$ will denote a model (in most cases a neural network); $x \in \mathcal{X}$ will be used to denote the set of features describing a given instance, which are used by $f$ to make a prediction $y \in \mathcal{Y}$. 

\subsection{Feature attribution methods}

\begin{figure}[t]
    \centering
    \includegraphics[width=0.5\textwidth]{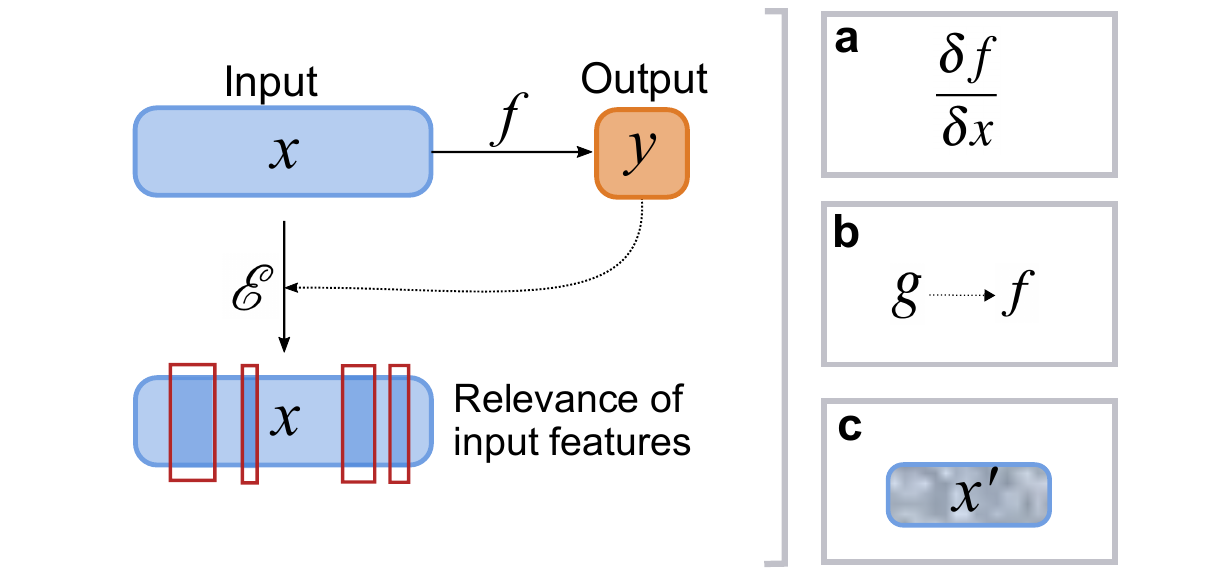}
    \caption{Feature attribution methods. Given a neural network model $f$, which computes the prediction $y$ = \textit{f(x)} for input sample $x$, a feature attribution method $\mathcal{E}$ outputs the relevance of every input feature of $x$ for the prediction. There are three basic approaches to determine feature relevance: (a) gradient-based methods, computing the gradient of the network $f$ \textit{w.r.t.} the input $x$, (b) surrogate methods, which approximate $f$ with a human-interpretable model $g$, and (c) perturbation-based methods, which modify the original input to measure the respective changes in the output.}
    \label{fig:featureattr}
\end{figure}

Given a regression or classification model $f: \boldsymbol{x} \in \mathbb{R}^K \rightarrow \mathbb{R}$, a feature attribution method is a function $\mathcal{E}:\boldsymbol{x}\in \mathbb{R}^K \rightarrow \mathbb{R}^K$ that takes the model input and produces an output whose values denote the relevance of every input feature for the final prediction computed with $f$. Feature attribution methods can be grouped into the following three categories (Figure~\ref{fig:featureattr}):

\begin{itemize}
    \item \textit{Gradient-based feature attribution}. These approaches measure how much a change around a local neighborhood of the input $\boldsymbol{x}$ corresponds to a change in the output $f(\boldsymbol{x})$. A common approach among deep-learning practitioners relies on the use of the derivative of the output of the neural network with respect to the input (\textit{i.e.}, $\dfrac{\delta f}{\delta \boldsymbol{x}}$) to determine feature importance \cite{sundararajan2017axiomatic, smilkov2017smoothgrad, adebayo2018local}. Its popularity arises partially from the fact that this computation can be performed via backpropagation \cite{rumelhart1986learning}, the main way of computing partial first-order derivatives in neural network models. While the use of gradient-based feature attribution may seem straightforward, several methods relying on this principle have been shown to lead to only partial reconstruction of the original features \cite{adebayo2018sanity}, which is prone to misinterpretation. 
    \item \textit{Surrogate-model feature attribution.} Given a model $f$, these methods aim to develop a surrogate explanatory model $g$, which is constructed in such a way that: (i) $g$ is interpretable, and (ii) $g$ approximates the original function $f$. A prominent example of this concept is the family of additive feature attribution methods, where the approximation is achieved through a linear combination of binary variables $z_i$:

\begin{equation}
    g\left(z^{'}_{i}\right) = \phi_0 + \sum_{i=1}^M \phi_i z_i,
\end{equation}

\noindent where $z_{i} \in \left\lbrace 0, 1 \right\rbrace ^ M$, $M$ is the number of original input features, $\phi_i \in \mathbb{R}$ are coefficients representing the importance assigned to each \textit{i}-th binary variable and $\phi_0$ is an intercept. Several notable feature attribution methods belong to this family \cite{lipovetsky2001analysis, lundberg2017unified}, such as Local Interpretable Model-Agnostic Explanations (LIME) \cite{ribeiro2016should}, Deep Learning Important FeaTures (DeepLIFT) \cite{shrikumar2017learning}, Shapley Additive Explanations (SHAP) \cite{lundberg2017unified} and Layer-Wise Relevance Propagation \cite{bach2015pixel}. Both gradient-based methods and the additive subfamily of surrogate attribution methods provide local explanations (\textit{i.e.}, each prediction needs to be examined individually), but they do not offer a general understanding of the underlying model $f$. {Global} surrogate explanation models aim to fill this gap by generically describing $f$ via a decision tree or decision set \cite{lakkaraju2017interpretable} model. If such an approximation is precise enough, these aim to to mirror the computation logic of the original model. While early attempts limited $f$ to the family of tree-based ensemble methods (\textit{e.g.}, random forests \cite{deng2019interpreting}), more recent approaches are readily applicable to arbitrary deep learning models \cite{bastani2017interpreting}.

\item\textit{Perturbation-based methods} modify or remove parts of the input aiming to measure its corresponding change in the model output; this information is then used to assess the feature importance. Alongside the well-established step-wise approaches \cite{maier1996use, balls1996investigating, sung1998ranking}, methods such as feature masking \cite{vstrumbelj2009explaining}, perturbation analysis \cite{fong2017interpretable}, response randomization \cite{olden2002illuminating}, and conditional multivariate models \cite{zintgraf2017visualizing} belong to this category. While perturbation-based methods have the advantage of directly estimating feature importance, they are computationally slow when the number of input features increases \cite{zintgraf2017visualizing}, and the final result tends to be strongly influenced by the number of features that are perturbed altogether \cite{ancona2017towards}.

\end{itemize}

Feature attribution methods have been used for ligand- and structure-based drug-discovery. For instance, McCloskey \textit{et al.} \cite{mccloskey2019using} employed gradient-based attribution \cite{sundararajan2017axiomatic} to detect ligand pharmacophores relevant for binding. The study showed that, despite good performance of the models on held-out data, these still can learn spurious correlations \cite{mccloskey2019using}. Pope \textit{et al.} \cite{pope2019explainability} adapted gradient-based feature attribution  \cite{selvaraju2017grad, zhang2018top} for the identification of relevant functional groups in adverse effect prediction \cite{tice2013improving}. Recently, SHAP \cite{lundberg2017unified} was used to interpret relevant features for compound potency and multi‑target activity prediction \cite{rodriguez2019interpretation, rodriguez2020interpretation}. Hochuli \textit{et al.} \cite{hochuli2018visualizing} compared several feature attribution methodologies, showing how the visualization of attributions assists in parsing and interpreting of protein-ligand scoring with 3D convolutional neural networks \cite{jimenez2018k, jimenez2019deltadelta}. 

It should be noted that the interpretability of feature attribution methods is limited by the original set of features (model input). In drug discovery the interpretability of machine learning methods is often hampered by the use of complex input molecular descriptors. When making use of feature attribution approaches, it is advisable to choose molecular descriptors or representations for model construction that one considers to bear `interpretable meaning' (Box 1).

\subsection{Instance-based approaches}

Instance-based approaches compute a subset of relevant features (instances) that must be present to retain (or absent to change) the prediction of a given model (Figure~\ref{fig:instancebased}). An instance can be real (\textit{i.e.,}  drawn from the set of data) or generated for the purposes of the method. Instance-based approaches have been argued to provide `natural' model interpretations for humans, because they resemble counterfactual reasoning (\textit{i.e.} producing alternative sets of action to achieve a similar or different result) \cite{doshi2017accountability}.

\begin{itemize}
    \item \textit{Anchor} algorithms \cite{ribeiro2018anchors} offer model-agnostic interpretable explanations of classifier models. They compute a subset of \textit{if}-\textit{then} rules based on one or more features that represent conditions to sufficiently guarantee a certain class prediction. In contrast to many other local explanation methods \cite{ribeiro2016should}, anchors therefore explicitly model the `coverage' of an explanation. Formally, an anchor $A$ is defined as a set of rules such that, given a set of features $x$ from a sample, they return $A(x) = 1$ if said rules are met, while guaranteeing the desired predicted class from $f$ with a certain probability $\tau$:

    \begin{equation}
    \mathbb{E}_{\mathcal{D}(z|A)}\left[ \mathbbm{1}_{f(x)=f(z)} \right] \geq \tau,
    \end{equation}

    \noindent where $\mathcal{D}(z|A)$ is defined as the conditional distribution on samples where anchor $A$ applies. This methodology has successfully been applied in several tasks such as image recognition, text classification, and visual question answering \cite{ribeiro2018anchors}.

\item \textit{Counterfactual instance search.} Given a classifier model $f$ and an original data point $x$, counterfactual instance search \cite{wachter2017counterfactual} aims to find examples $x'$ that (i) are as close to $x$ as possible, and (ii) for which the classifier produces a different class label from the label assigned to $x$. In other words, a counterfactual describes small feature changes in sample $x$ such that it is classified differently by $f$. The search for the set of instances $x'$  may be cast into an optimization problem:

\begin{equation}
    \min_{x'} \max_{\lambda} \left(f_t - p_t \right)^2 + \lambda L_1 \left(x', x \right),
\end{equation}

\noindent where $f_t$ is the prediction of the model for the $t$-th class, $p_t$ a user-defined target probability for the same class, $L_1$ the Manhattan distance between the proposed $x'$ and the original sample $x$, and $\lambda$ an optimizable parameter that controls the contribution of each term in the loss. The first term in this loss encourages the search towards points that change the prediction of the model, while the second ensures that both $x$ and $x'$ lie close to each other in their input manifold. While in the original paper this approach was shown to successfully obtain counterfactuals in several data sets, the results revealed a tendency to look artificial. A recent methodology \cite{van2019interpretable} mitigates this problem by adding extra terms to the loss function with an autoencoder architecture \cite{kramer1991nonlinear}, to better capture the original data distribution. Importantly, counterfactual instances can be evaluated using trust scores (\textit{cf}. the section on uncertainty estimation). One can interpret a high trust score as the counterfactual being far from the initially predicted class of $x$ compared to the class assigned to the counterfactual $x'$.

\begin{figure}[t]
    \centering
    \includegraphics[width=0.5\textwidth]{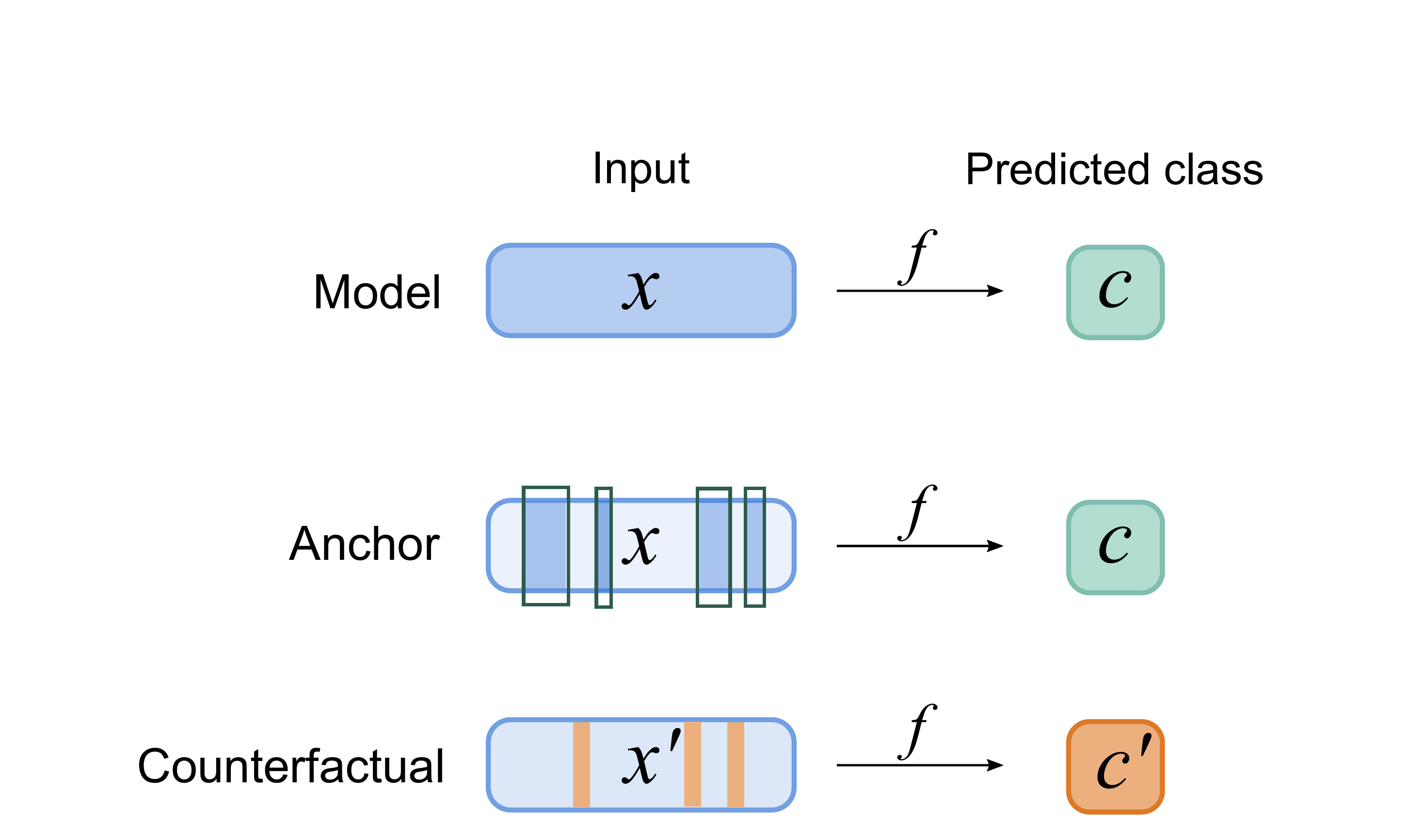} % v2 has colors for classes, v3 doesn't
    \caption{Instance-based model interpretation. Given a model $f$, input instance $x$ and the respective predicted class $c$, so-called anchor algorithms identify a minimal subset of features of $x$ that are sufficient to preserve the predicted class assignment $c$. Counterfactual search generates a new instance $x'$ that lies close in feature space to $x$ but is classified differently by the model, as belonging to class $c'$.}
    \label{fig:instancebased}
\end{figure}

\item \textit{Contrastive explanation} methods \cite{dhurandhar2018explanations} provide instance-based interpretability of classifiers by generating `pertinent positive' (PP) and `pertinent negative' (PN) sets. This methodology is, therefore, related to both anchors and counterfactual search approaches. PPs are defined as the smallest set of features that should be present in an instance in order for the model to predict a `positive' result (similar to anchors). Conversely, PNs identify the smallest set of features that should be absent in order for the model to be able to sufficiently differentiate from the other classes (similar to a counterfactual instance). This method generates explanations of the form \textit{"An input $x$ is classified as class $y$ because a subset of features $x_1, \dots, x_k$ is present, and because a subset of features $x_m, \dots, x_p$ is absent." \cite{doshi2017accountability, herman2016you}}. Contrastive explanation methods find such sets by solving two separate optimization problems, namely by (i) perturbing the original instance until it is predicted differently than its original class, and (ii) searching for critical features in the original input (\textit{i.e.,} those features that guarantee a prediction with a high degree of certainty). The proposed approach uses an elastic net regularizer \cite{zou2005regularization}, and optionally a conditional autoencoder model \cite{mousavi2017deepcodec} so that the found explanations are more likely to lie closer to the original data manifold.
\end{itemize}

To the best of our knowledge, instance-based approaches have yet to be applied to drug-discovery. In our opinion, they bear promise in several areas of de novo molecular design, such as (i) activity cliff prediction \cite{stumpfe2012exploring, cruz2014activity}, as they can help identify small structural variations in molecules that cause large bioactivity changes, (ii) fragment-based virtual screening \cite{erlanson2011introduction}, by highlighting a minimal subset of atoms responsible for a given observed activity), and (iii) hit-to-lead optimization, by helping identify the minimal set of structural changes required to improve one or more biological or physicochemical properties.

\subsection{Graph-convolution-based methods}

Molecular graphs are a natural mathematical representation of molecular topology, with nodes and edges representing atoms and chemical bonds, respectively \cite{todeschini2009molecular} (Figure~\ref{fig:graph}a). Furthermore, their usage has been commonplace in chemoinformatics and mathematical chemistry since the late 1970s \cite{randic1979search, kier1985shape, randic1975unique, bonchev1977information}. Thus, it does not come as a surprise in these fields to witness the increasing application of novel graph-convolution neural networks \cite{duvenaud2015convolutional}, which formally fall under the umbrella of neural-message passing algorithms \cite{gilmer2017neural, kipf2016semi, nguyen2017semi}. Generally speaking, convolution refers to a mathematical operation on two functions that produces a third function expressing how the shape of one is modified by the other. This concept is widely used in convolutional neural networks for image analysis. Graph convolutions naturally extend the convolution operation typically used in computer vision \cite{krizhevsky2012imagenet} or in natural language processing \cite{kim2014convolutional} applications to arbitrarily-sized graphs. In the context of drug discovery, graph convolutions have been applied to molecular property prediction \cite{kearnes2016molecular, wu2018moleculenet} and in generative models for de novo drug design \cite{jin2018junction}.

Exploring the interpretability of models trained with graph-convolution architectures is currently a particularly active research topic. For the purpose of this review, XAI methods based on graph-convolution are grouped into the following two categories:

\begin{itemize}
    \item \textit{Sub-graph identification approaches} aim to identify one or more parts of a graph that are responsible for a given prediction (Figure~\ref{fig:graph}b). GNNExplainer \cite{ying2019gnnexplainer} is a model-agnostic example of such category, and provides explanations for any graph-based machine learning task. Given an individual input graph, GNNExplainer identifies a connected sub-graph structure, as well a set of node-level features that are relevant for a particular prediction. The method can also provide such explanations for a group of data points belonging to the same class. GNNExplainer is formulated as an optimization problem, where a mutual information objective between the prediction of a graph neural network and the distribution of feasible sub-graphs is maximized.  Mathematically, given a node $v$, the goal is to identify a sub-graph $G_S \subseteq	G$ with associated features $X_S = \left\lbrace x_j | v_j \in G_S \right\rbrace$ that are relevant in explaining a target prediction $\hat{y} \in Y$ via a mutual information measure $\text{MI}$:

\begin{multline}
    \max_{G_S} \text{MI} \left(Y, \left( G_S, X_S \right)  \right) = H(Y) - \\ H(Y|G = G_S, X=X_S).
\end{multline}

In practice, however, this objective is not mathematically tractable, and several continuity and convexity assumptions have to be made. GNNExplainer was tested on a set of molecules labeled for their mutagenic effect on \textit{Salmonella typhimurium} \cite{debnath1991structure}, and identified several known mutagenic functional groups (\textit{i.e.}, certain aromatic and heteroaromatic rings and amino/nitro groups \cite{debnath1991structure}) as relevant. 

\item \textit{Attention-based approaches}. The interpretation of graph-convolutional neural networks can benefit from attention mechanisms \cite{velivckovic2017graph}, which borrow from the natural language processing field, where their usage has become standard \cite{bahdanau2014neural}. The idea is to stack several message-passing layers to obtain hidden node-level representations, by first computing attention coefficients associated with each of the edges connected to the neighbors of a particular node in the graph (Figure~\ref{fig:graph}c). Mathematically, for a given node, an attention-based graph convolution operation obtains its updated hidden representation via a normalized sum of the node-level hidden features of the topological neighbours:

\begin{equation}
    h_i^{\left(l +1 \right)} = \sigma \left( \sum_{j \in \mathcal{N}(i)} \alpha_{ij}^{l} W^{\left(l \right)} h_j^{\left(l \right)} \right),
    \label{eq:attention}
\end{equation}

\begin{figure}[t]
    \centering
    \includegraphics[width=0.5\textwidth]{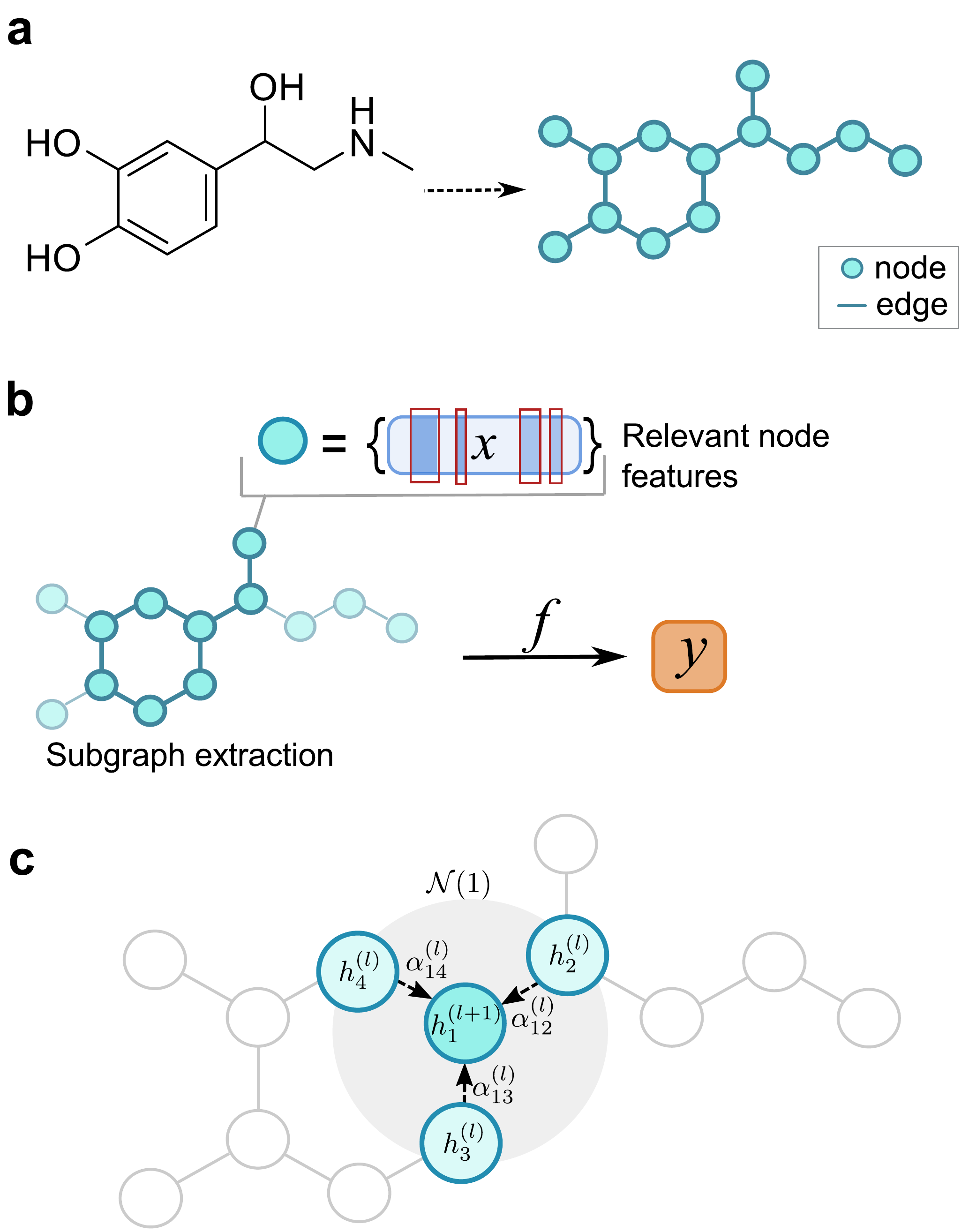}
    \caption{Graph-based model interpretation. (a) Kekul\'{e} structure of adrenaline and its respective molecular graph; atoms and bonds constitute nodes and edges, respectively. (b) Given an input graph, approaches like GNNExplainer \cite{ying2019gnnexplainer} aim to identify a connected, compact subgraph, as well as node-level features that are relevant for a particular prediction $y$ of a graph-neural network model $f$. (c) Attention mechanisms can be used in conjunction with message-passing algorithms in order to learn coefficients $\alpha ^{(l)}_{ij}$ for the $l$-th layer, which assign 'importance' to the set of neighbours $\mathcal{N}(i)$ (\textit{e.g.}, adjacent atoms) of a node $i$. These coefficients are an explicit component in the computation of new hidden-node representations $h_{i}^{(l+1)}$ (Eq.~\eqref{eq:attention}) in attention-based graph-convolutional architectures. Such learned attention coefficients can be then used to highlight the predictive relevance of certain edges and nodes.}
    \label{fig:graph}
\end{figure}

\noindent where $\mathcal{N}(i)$ is the set of topological neighbours of node \textit{i} with a one-edge distance, $\alpha_{ij}^{l}$ are learned attention coefficients over those neighbours, $\sigma$ is a non-linear activation function, and $W^{\left(l \right)}$ is a learnable feature matrix for layer $l$. The main difference between this approach and a standard graph convolution update is that, in the latter, attention coefficients are replaced by a fixed normalization constant $c_{ij} = \sqrt{|\mathcal{N}(i)|} \sqrt{|\mathcal{N}(j)|}$.

\end{itemize}

A recent study \cite{preuer2019interpretable} describes how the interpretation of filters in message-passing networks can lead to the identification of relevant pharmacophore- and toxicophore-like substructures, showing consistent findings with literature reports. Gradient-based feature attribution techniques, such as Integrated Gradients \cite{sundararajan2017axiomatic}, were used in conjunction with graph-convolutional networks to analyze retrosynthetic reaction predictions and highlight the atoms involved in each reaction step \cite{ishida2019prediction}. Attention-based graph convolutional neural networks have also been used for the prediction of solubility, polarity, synthetic accessibility, and photovoltaic efficiency, among other properties \cite{shang2018edge, ryu2018deeply}, leading to the identification of relevant molecular substructures for the target properties. Finally, attention-based graph architectures have also been used in chemical reactivity prediction \cite{coley2019graph}, pointing to structural motifs that are consistent with a chemist's intuition in the identification of suitable reaction partners and activating reagents. 

Due to their intuitive connection with the two-dimensional representation of molecules, graph-convolution-based XAI bears the potential of being applicable in many common modeling tasks in drug discovery. Some of the applications that might benefit most from the usage of these approaches are those aiming to identify relevant structural molecular motifs, \textit{e.g.,} for structural alert identification \cite{limban201structalert}, site of reactivity \cite{hughes2015site} or metabolism \cite{kirchmair2012computational} prediction, and the selective optimization of side activities \cite{wermuth2006selective}.

\subsection{Self-explaining approaches}

The XAI methods introduced so far produce \textit{a posteriori} explanations of deep learning models. Although such \textit{post-hoc} interpretations have been shown to be useful, some argue that, ideally, XAI methods, should automatically offer human-interpretable explanation alongside their predictions \cite{laugel2019dangers}. Such approaches (herein referred to as  `\textit{self-explaining}') would promote verification and error analysis, and be directly linkable with domain knowledge \cite{melis2018towards}. While the term self-explaining has been coined to refer to a specific neural network architecture -- self-explaining neural networks \cite{melis2018towards}, described below -- in this review, the term is used in a broader sense, so as to identify methods that feature interpretability as a central part of their design. Self-explaining XAI approaches can be grouped into the following categories:

\begin{itemize}
    \item \textit{Prototype-based reasoning} refers to the task of forecasting future events (\textit{i.e.}, novel samples) based on particularly informative known data points. Usually, this is done by identifying \textit{prototypes}, \textit{i.e.}, representative samples, which are adapted (or used directly) to make a prediction. These methods are motivated by the fact that predictions based on individual, previously seen examples mimic human decision making \cite{leake1996case}. The Bayesian case model \cite{kim2014bayesian}, is a \textit{pre}-deep-learning approach that constitutes a general framework for such prototype-based reasoning. A Bayesian case model learns to identify observations that best represent clusters in a dataset (\textit{i.e.}, prototypes), along with a set of defining features for that cluster. Joint inference is performed on cluster labels, prototypes, and extracted relevant features, thereby providing interpretability without sacrificing classification accuracy \cite{kim2014bayesian}. Recently, Li \textit{et al.} \cite{li2018deep} developed a neural network architecture composed of an autoencoder and a therein named `prototype layer', whose units store a learnable weight vector representing an encoded training input. Distances between the encoded latent space of new inputs and the learned prototypes are then used as part of the prediction process. This approach was later expanded by Chen \textit{et al.} \cite{chen2019looks} to convolutional neural networks for computer vision tasks.

    \item \textit{Self-explaining neural networks} \cite{melis2018towards} aim to associate input or latent features with semantic concepts. They jointly learn a class prediction and generate explanations using a feature-to-concept mapping. Such a network model consists of (i) a sub-network that maps raw inputs into a predefined set of explainable concepts, (ii) a parameterizer that obtains coefficients for each individual explainable concept, and (iii) an aggregation function that combines the output of the previous two components to produce the final class prediction. 
    
    \item \textit{Human-interpretable concept learning} refers to the task of learning a class of concepts, \textit{i.e.} high-level combinations of knowledge elements \cite{goodman2014concepts}) from data, aiming to achieve human-like generalization ability. The Bayesian Program Learning approach \cite{lake2015human} was proposed with the goal of learning visual concepts in computer vision tasks. Such concepts were represented as probabilistic programs expressed as structured procedures in an abstract description language \cite{ghahramani2015probabilistic}. The model then composes more complex programs using the elements of previously learned ones using a Bayesian criterion. This approach was shown to reach human-like performance in one-shot learning tasks \cite{vinyals2016matching, altae2017low}.
    
    \item \textit{Testing with concept activation vectors} \cite{kim2017interpretability} computes the directional derivatives of the activations of a layer \textit{w.r.t.} its input, towards the direction of a concept. Such derivatives quantify the degree to which the latter is relevant for a particular classification (\textit{e.g.}, how important the concept `stripes' is for the prediction of the class `zebra'). It does so by considering the mathematical association between the internal state of a machine learning model -- seen as a vector space $E_m$ spanned by basis vectors $e_m$ that correspond to neural activations -- and human-interpretable activations residing in a different vector space $E_h$ spanned by basis vectors $e_h$. A linear function is computed that translates between these vector spaces ($g: E_m \rightarrow E_h$). The association is achieved by defining a vector in the direction of the values of a concept's set of examples, and then training a linear classifier between those and random counterexamples, to finally take the vector orthogonal to the decision boundary.
    
    \item \textit{Natural language explanation generation}.
    Deep networks can be designed to generate human-understandable explanations in a supervised manner \cite{gilpin2018explaining}. In addition to minimizing the loss of the main modeling task, several approaches synthesize a sentence using natural language that explains the decision performed by the model, by simultaneously training generators on large data sets of human-written explanations. This approach has been applied to generate explanations that are both image- and class-relevant \cite{hendricks2016generating}. Another prominent application is visual question answering \cite{antol2015vqa}. To obtain meaningful explanations, however, this approach requires a significant amount of human-curated explanations for training, and might, thus, find limited applicability in drug discovery tasks.
\end{itemize}

To the best of our knowledge, self-explaining deep learning has not been applied to chemistry or drug design yet. Including interpretability by design could help bridge the gap between machine representation and the human understanding of many types of problems in drug discovery. For instance, prototype-reasoning bears promise in the modeling of heterogeneous sets of chemicals with different modes of action, allowing to preserve both mechanistic interpretability and predictive accuracy. Explanation generation (either concept- or text-based) is another potential solution to include human-like reasoning and domain knowledge in the model building task. In particular, explanation generation approaches might be applicable to certain decision-making processes, such as the replacement of animal testing and \textit{in-vitro} to \textit{in-vivo} extrapolation, where human-understandable generated explanations constitute a crucial element.

\subsection{Uncertainty estimation}

\begin{figure}[t]
    \centering
    \includegraphics[width=0.4\textwidth]{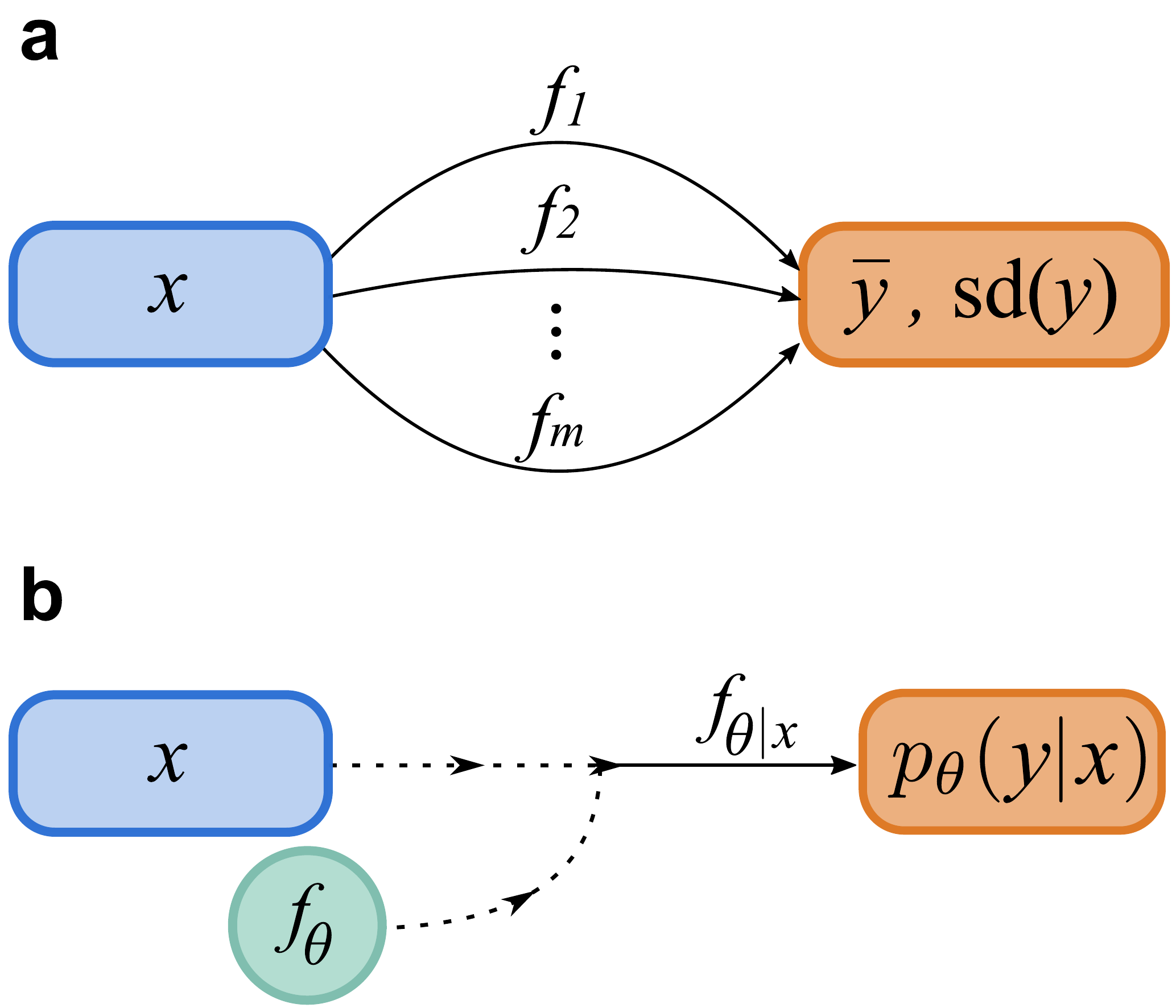}
    \caption{Uncertainty estimation. (a) Ensemble-based methods aggregate the output of $m$ identical, but differently-initialized, models $f_i$. The final prediction is obtained by aggregating the predictions of all models (\textit{e.g.,} as the average, $\bar{y}$), while an uncertainty estimate can be obtained from the respective predictive variance, \textit{e.g.,} in the form of a standard deviation, $\mathrm{sd}(y)$. (b) Bayesian probabilistic approaches consider a prior $p(\theta)$ over the learnable weights of a neural network model $f_\theta$, and make use of approximate sampling approaches to learn a posterior distribution over both the weights $p(\theta | x)$ and the prediction $p_\theta(y | x)$. These distributions can be then sampled from to obtain uncertainty estimates over both the weights and the predictions.}
    \label{fig:uncertainty}
\end{figure}

Uncertainty estimation,\textit{ i.e.} the quantification of epistemic error in a prediction, constitutes another approach to model interpretation. While some machine learning algorithms, such as Gaussian processes \cite{rasmussen2003gaussian}, provide built-in uncertainty estimation, deep neural networks are known for being poor at quantifying uncertainty \cite{nguyen2015deep}. This is one of the reasons why several efforts have been devoted to specifically quantify uncertainty in neural network-based predictions. Uncertainty estimation methods can be grouped into the following categories:

\begin{itemize}
    \item \textit{Ensemble approaches}. Model ensembles improve the overall prediction quality and have become a standard for uncertainty estimates \cite{hansen1990neural}. Deep ensemble averaging \cite{lakshminarayanan2017simple} is based on $m$ identical neural network models that are trained on the same data and with a different initialization. The final prediction is obtained by aggregating the predictions of all models (\textit{e.g.,} by averaging), while an uncertainty estimate can be obtained from the respective variance (Figure ~\ref{fig:uncertainty}a). Similarly, the sets of data on which these models are trained can be generated via bootstrap re-sampling \cite{freedman1981bootstrapping}. A disadvantage of this approach is its computational demand, as the underlying methods build on $m$ independently trained models. Snapshot ensembling \cite{huang2017snapshot} aims to overcome this limitation by periodically storing model states (\textit{i.e.} model parameters) along the the training optimization path. These model `snapshots' can be then used for constructing the ensemble.

    \item \textit{Probabilistic approaches} aim to estimate the posterior probability of a certain model output or to perform \textit{post-hoc} calibration. Many of these methods treat neural networks as Bayesian models, by considering a prior distribution over its learnable weights, and then performing inference over their posterior distribution with various methods (Figure ~\ref{fig:uncertainty}b), \textit{e.g.,} Markov Chain Monte Carlo \cite{zhang2019cyclical} or variational inference \cite{graves2011practical, sun2019functional}. Gal \textit{et al.} \cite{gal2016dropout}, suggested the usage of dropout regularization to perform approximate Bayesian inference, which was later extended\cite{kendall2017uncertainties} to compute epistemic (\textit{i.e.}, caused by model mis-specification) and aleatory uncertainty (inherent to the noise in the data). Similar approximations can also be made via batch normalization \cite{teye2018bayesian}. Mean variance estimation \cite{nix1994estimating}  considers a neural network designed to output both a mean and variance value, to then train the model using a negative Gaussian log-likelihood loss function. Another subcategory of approaches consider asymptotic approximations of a prediction by making Gaussian distributional assumptions of its error, such as the delta technique \cite{chryssolouris1996confidence, hwang1997prediction}.
    \item \textit{Other approaches}. The LUBE (lower upper bound estimation) \cite{khosravi2010lower} approach trains a neural network with two outputs, corresponding to the upper and lower bounds of the prediction. Instead of quantifying the error of single predictions, LUBE uses simulated annealing and optimizes the model coefficients to achieve (a) maximum coverage (probability that the real value of the \textit{i-}th sample will fall between the upper and the lower bound) of training measurements and (b) minimum prediction interval width.  Ak \textit{et al.} suggested to quantify the uncertainty in neural network models by directly modelling interval-valued data \cite{ak2015interval}. Trust scores \cite{jiang2018trust} measure the agreement between a neural network and a $k$-nearest neighbour classifier that is trained on a filtered subset of the original data. The trust score considers both the distance between the instance of interest to the nearest class that is different from the original predicted one and its distance towards the predicted class. Union-based methods \cite{huang2015scalable} first train a neural network model and then feed its embeddings to a second model that handles uncertainty, such as a Gaussian process or a random forest. Distance-based approaches \cite{sheridan2004similarity} aim to estimate the prediction uncertainty of a new sample $x'$ by measuring the distance to the closest sample in the training set, either using input features \cite{liu2018molecular} or an embedding produced by the model \cite{janet2019quantitative}. 
\end{itemize}

Many uncertainty estimation approaches have been successfully implemented in drug discovery applications \cite{scalia2020evaluating}, mostly in traditional QSAR modeling, either by the use of models that naturally handle uncertainty \cite{obrezanova2007gaussian} or post-hoc methods \cite{schroeter2007estimating, clark2014using, bosc2019large}. Attention has recently been drawn toward the development of uncertainty-aware deep-learning applications in the field. 'Snapshot ensembling' was applied to model 24 bioactivity datasets \cite{cortes2018deep}, showing that it performs \textit{on par} with random forest and neural network ensembles, and also leads to narrower confidence intervals. Schwaller \textit{et al.} \cite{schwaller2019molecular} proposed a transformer model \cite{vaswani2017attention} for the task of forward chemical reaction prediction. This approach implements uncertainty estimation by computing the product of the probabilities of all predicted tokens in a SMILES sequence representing a molecule. Zhang \textit{et al.} \cite{zhang2019bayesian} have recently proposed a Bayesian treatment of a semi-supervised graph-neural network for uncertainty-calibrated predictions of  molecular properties, such as the melting point and aqueous solubility. Their results suggest that this approach can efficiently drive an active learning cycle, particularly in the low-data regime -- by choosing those molecules with the largest estimated epistemic uncertainty. 

Importantly, a recent comparison of several uncertainty estimation methods for physicochemical property prediction showed that none of the methods systematically outperformed all others \cite{hirschfeld2020uncertainty}.

\section{Available software}

Given the attention deep learning applications are currently receiving, several software tools have been developed to facilitate model interpretation. A prominent example is Captum \cite{captum2019github}, an extension of the PyTorch \cite{paszke2019pytorch} deep-learning and automatic differentiation package that provides support for most of the feature attribution techniques described in this work. Another popular package is Alibi \cite{alibi2020}, which provides instance-specific explanations for certain models trained with the scikit-learn \cite{scikit-learn} or TensorFlow \cite{abadi2016tensorflow} packages. Some of the explanation methods implemented include anchors, contrastive explanations, and counterfactual instances.

\section{Conclusions and outlook}

Automated analysis of medical and chemical knowledge to extract and present features in a human-readable format dates back to the 1990s \cite{hirst1994quantitative, fiore1995integrated}, but it's receiving increasing attention in the last years due to the \textit{renaissance} of neural networks in chemistry and healthcare. Given the current pace of AI in drug discovery and related fields, there will be an increased demand for methods that help us understand and interpret deep learning models. 

For drug discovery and design in particular, XAI will foster collaborations between medicinal chemists, chemoinformaticians and data scientists. If successful, XAI bears the potential to provide fundamental support in the analysis and interpretation of increasingly more complex chemical data, as well as in the formulation of new pharmacological hypotheses, while avoiding human biases \cite{kutchukian2012inside, boobier2017can}. At the same time, novel drug discovery challenges might boost the development of application-tailored XAI approaches, to promptly respond to specific scientific questions related to human biology and pathophysiology. 

It will be important to explore the opportunities and limitations of the established chemical language for representing the decision space of these models. Additionally, we have to concede our incomplete understanding of the human pathology at the molecular level, with all its individual idiosyncrasies. In this context, full comprehensibility may be hard to achieve, given, for instance, the often nonlinear relationships between chemical structure and pharmacological activity, the presence of error, and limited predictability \cite{schneider2016novo}.

XAI also poses technical challenges, given the multiplicity of possible explanations and methods applicable to a given task \cite{lipton2017doctor}. Most approaches do not come as readily usable out-of-the-box solutions, but need to be tailored to each individual application. Additionally, profound knowledge of the problem domain is crucial to identify which model decisions demand further explanations, which type of answers are meaningful to the user and which are instead trivial or expected \cite{goodman2017european}. For human decision making, the explanations generated with XAI have to be \textit{non-trivial}, \textit{non-artificial}, and \textit{sufficiently informative} for the respective scientific community. At least for the time being, finding such solutions will require the collaborative efforts of deep-learning experts, chemoinformaticians and data scientists, chemists, biologists and other domain experts, to ensure that XAI methods serve their intended purpose and deliver reliable answers.

A crucial challenge for future AI-assisted drug discovery is the data representation used for machine and deep learning. In contrast to many other areas in which deep learning has been shown to excel, such as natural language processing and image recognition, there is no naturally applicable, complete, `raw' molecular representation. After all, molecules -- as scientists conceive them -- are models themselves.  Such `induction-based' approach, which builds higher-order (\textit{e.g.}, deep learning) models  from lower-order ones (\textit{e.g.}, molecular representations or descriptors based on observational statements) is therefore philosophically debatable \cite{bendassolli2013theory}. The question as to how to represent molecules for deep learning thus constitutes still one of the fundamental challenges of XAI in drug discovery and design. 

One step forward towards explainable AI is to build on interpretable `low-level' molecular representations that have direct meaning for chemists (\textit{e.g.} SMILES strings \cite{ikebata2017bayesian, segler2018generating}, amino acid sequences \cite{nagarajan2018computational, muller2018recurrent}, and spatial 3D-voxelized representations \cite{jimenez2018k, jimenez2020deep}). Many recent studies also rely on well-established molecular descriptors, such as hashed binary fingerprints \cite{rogers2010extended, awale2014atom} and topochemical and geometrical descriptors \cite{todeschini2010new, katritzky1993traditional}, which capture structural features defined \textit{a priori}. Often, molecular descriptors, while being relevant for subsequent modeling, capture complex chemical information in a non readily-interpretable way. Consequently, when striving for XAI, there is an understandable tendency to employ molecular representations that are more easily interpretable in terms of the known language of chemistry \cite{vighi2019predictive}. It therefore goes without saying that the development of novel interpretable molecular representations for deep learning will constitute a critical area of research for the years to come, including the development of self-explaining approaches to overcome the hurdles of non-interpretable but information-rich descriptors, by providing human-like explanations alongside sufficiently accurate predictions.

Most of the deep learning models in drug discovery do not consider applicability domain issues \cite{sahigara2012comparison}, \textit{i.e.}, the region of chemical space where statistical learning assumptions are met \cite{mathea2016chemoinformatic}. Inclusion of applicability domain restrictions -- in an interpretable way, whenever possible -- should be considered as an integral element of explainable AI. Knowing when to apply which particular model will, in fact, help address the problem of high confidence of deep learning models on wrong predictions \cite{nguyen2015deep, szegedy2013intriguing, hendrycks2016baseline}, avoiding unnecessary extrapolations at the same time. Assessing a model's applicability domain and rigorous determination of model accuracy might arguably be more important for decision making than the particular modeling approach chosen \cite{liu2019ad}.

At present, XAI in drug discovery is lacking open-community platforms, where software code, model interpretations, and the respective training data can be shared and improved by synergistic efforts of researchers with different scientific backgrounds. Initiatives like MELLODDY (\textit{Machine Learning Ledger Orchestration for Drug Discovery}) for decentralized, federated model development and secure data handling across pharmaceutical companies constitute a first step in the right direction \cite{hale2019melloddy}. Such kinds of collaboration will hopefully foster the validation and acceptance of XAI approaches and the associated explanations they provide. 

Especially in time- and cost-sensitive scenarios like drug discovery, the users of deep learning methods have the responsibility to cautiously inspect and interpret the predictions made by such a computational model. Keeping in mind the possibilities and limitations of drug discovery with XAI, it is reasonable to assume that the continued development of alternative models that are more easily comprehensible and computationally affordable will not lose its importance.

{\footnotesize

\section*{Conflict of interest}
G.S. declares a potential financial conflict of interest in his role as a co-founder of inSili.com GmbH, Zurich, and consultant to the pharmaceutical industry.

\section*{Author contributions}
All authors contributed equally to this manuscript.

\section*{Acknowledgements}
Dr. Nils Weskamp is thanked for helpful feedback on the manuscript. This work was financially supported by the ETH RETHINK initiative, the Swiss National Science Foundation (grant no. 205321${\_}$182176), and Boehringer Ingelheim Pharma GmbH \& Co. KG.

\section*{Related links}

\begin{itemize}
    \item PyTorch Captum: \url{captum.ai}
    \item Alibi: \url{github.com/SeldonIO/alibi}
    \item MELLODDY Consortium: \url{melloddy.eu}
\end{itemize}

\bibliography{references}{}
\bibliographystyle{naturemag}
% {\footnotesize \putbib}
\clearpage
}

\pagebreak
\begin{tcolorbox}[width=1\textwidth, colback=white]
\textbf{Box 1: Explainable AI (XAI) applied to cytochrome P450 metabolism.}

\vspace{5mm}
This worked example showcases XAI that provides a graphical explanation in terms of molecular motifs that are considered relevant by a neural network model predicting drug interaction with cytochrome P450 (3A4 isoform, CYP3A4).  The \textit{integrated gradients feature attribution} method \cite{sundararajan2017axiomatic} was combined with a graph-convolutional neural network for predicting drug-CYP3A4 interaction. This network model was trained with a publicly available set of CYP3A4 substrates and inhibitors \cite{nembri2016silico}. The figure shows the results obtained for two drugs that are metabolized predominantly by CYP3A4, namely the phosphodiesterase A inhibitor (antiplatelet agent)  cilostazol \cite{hiratsuka2007characterization}, and nifedipine,\cite{foti2010selection} an L-type calcium channel blocker. 

\begin{center}
\includegraphics[width=1\textwidth]{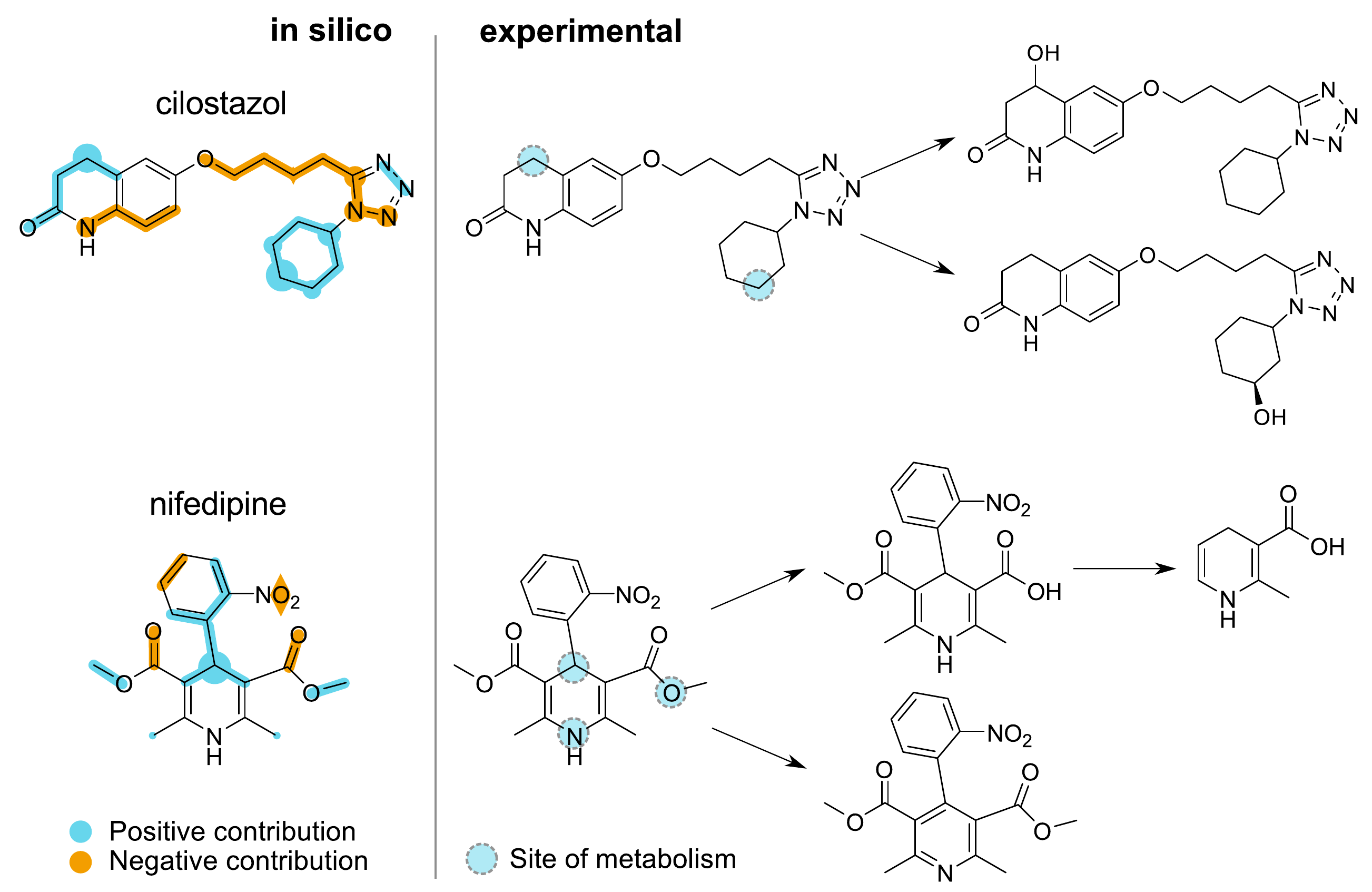}
\end{center}

The structural features for CYP3A4-compound interaction  suggested by the XAI method are highlighted in color (\textit{left panel} `in silico': blue, positive contribution to interaction with CYP3A4; orange, negative contribution to interaction; spot size indicates the feature relevance of the respective atom). The  main sites of metabolism (dashed circles) and the known metabolites \cite{waller1984first, hiratsuka2007characterization, raemsch1983pharmacokinetics, wishart2018drugbank} are shown in the right panel (`experimental'). Apparently, the XAI  captured the chemical substructures involved in CYP3A4-mediated biotransformation and most of the known sites of metabolism. Additional generic features related to metabolism were identified, \textit{i.e.}, (i) the tetrazole moiety and the secondary amino group in cilostazol, which are known to increase metabolic stability (\textit{left panel}: orange, negative contribution to the CYP3A4 -cilostazol interaction), and (ii) metabolically labile groups, such as methyl and ester groups (\textit{left panel}: blue, positive contribution to the CYP3A4-nifedipine interaction).

\end{tcolorbox}

\clearpage

\end{document}